# Scientific and Technological Text Knowledge Extraction Method of based on Word Mixing and GRU

Suyu Ouyang, Yingxia Shao*, Junping Du and Ang Li

（*College of Computer Scienc, Beijing Key Laboratory of Intelligent Telecommunication Software and Multimedia, Beijing* 100082）

（darkham97@163.com）

**Abstract** The knowledge extraction task is to extract triple relations (head entity-relation-tail entity) from unstructured text data. The existing knowledge extraction methods are divided into "pipeline" method and joint extraction method. The "pipeline" method is to separate named entity recognition and entity relationship extraction and use their own modules to extract them. Although this method has better flexibility, the training speed is slow. The learning model of joint extraction is an end-to-end model implemented by neural network to realize entity recognition and relationship extraction at the same time, which can well preserve the association between entities and relationships, and convert the joint extraction of entities and relationships into a sequence annotation problem. In this paper, we propose a knowledge extraction method for scientific and technological resources based on word mixture and GRU, combined with word mixture vector mapping method and self-attention mechanism, to effectively improve the effect of text relationship extraction for Chinese scientific and technological resources.

## 1 Introduction

Knowledge extraction is not only one of the tasks of information extraction, but also a key step in constructing and completing knowledge graphs [1]. The knowledge extraction task [2] is to extract triple relations (head entity-relation-tail entity) from unstructured text data. The existing knowledge extraction methods are divided into "pipeline" method and joint extraction method. The "pipeline" method is to separate named entity recognition and entity knowledge extraction into separate modules. Although this method has better flexibility, the training speed is slow. The learning model of joint extraction is an end-to-end model realized by neural network to realize entity recognition and knowledge extraction at the same time, which can well preserve the association between entities and relationships, and convert the joint extraction of entities and relationships into a sequence annotation. question.

this paper proposes the joint knowledge extraction model, which well preserves the association between entities and relations, and transforms the joint extraction of entities and relations into a sequence labeling problem. In order to avoid boundary segmentation errors to the greatest extent, the method of word labeling is selected, that is, the input is performed with words as the basic unit. However, in Chinese, it is difficult to store effective semantic information with simple word Embedding. Therefore, in order to integrate semantic information more effectively, a word mixing method is designed. At the same time, the self-attention mechanism is combined to capture long-distance semantic information in sentences, and the model extraction effect is improved by introducing bias weights.

The main contributions of this paper include three aspects:

1) A knowledge extraction method for scientific and technological texts (MBGAB) based on word mixing and GRU is proposed, which combines the attention mechanism to extract the relationship between Chinese scientific and technological resource texts.

2) The vector mapping method of word-mixing is used to avoid boundary segmentation errors to the greatest extent, and at the same time effectively integrate semantic information.

3) An end-to-end joint extraction model is adopted,

This work is supported by National Key R&D Program of China (2018YFB1402600), the National Natural Science Foundation of China (61772083, 61877006, 61802028, 62002027)
**Corresponding author** : Yingxia Shao ( shaoyx@bupt.edu.cn)

a bidirectional GRU network is used, and a self-attention mechanism is used to effectively capture long-distance semantic information in sentences, and the model extraction effect is improved by introducing bias weights.

## 2 Related work

Whether it is a professional technology resource platform or a social media scene [3], there is a large amount of scientific and technological text data [4], and knowledge extraction of this information can be used for better mining[5] and utilization[6]. With the development of deep learning technology, the use of neural networks[7] to extract information has become a common practice. In recent years, long short-term memory network (LSTM) allows each neural unit to forget or retain information, which is mainly used to overcome the characteristic of recurrent neural network (RNN) that historical information is gradually forgotten as the sequence length increases. The Gated Recurrent Unit (GRU) [8] was originally designed to allow each recurrent unit to be adaptively tuned at different times to capture dependencies. This model is simpler than LSTM. Bahdanau D et al. [9] proposed an attention mechanism that utilizes all the hidden states of the encoder RNN to help the decoding process, mimicking that humans can focus on certain parts of a sentence.

Currently, entity and relation extraction mainly includes pipeline extraction and joint extraction. Pipeline extraction usually separates named entity recognition and semantic relation classification. For the named entity recognition task, deep learning transforms it into a sequence labeling task. Lafferty J et al. proposed Conditional Random Field (CRF) [10], which combined the characteristics of maximum entropy model and hidden Markov model [11][12], and achieved good results in sequence tagging tasks such as part-of-speech tagging and named entity recognition. Collobert et al. [13] employed a combined CNN and CRF network to encode the word embedding layer. The task of semantic relation classification has made progress in recent years, and the most widely used are convolutional neural network (CNN)[14][15][16], recurrent neural network (RNN) [17][18][19] and long short-term memory network (LSTM) [20][21]. In addition, there are other methods, the combined model FCM [22] and the semantic learning model [23][24] can learn substructure representations of annotated sentences, which can handle global information input of any information and combination type.

Liberature [25] combined a recurrent neural network with a convolutional neural network, and the shared layers mainly shared the word embeddings and implicitly encoded information of the two. The network classifies semantic relations. Miwa et al. [26] also adopted a similar approach, superimposing a bi-directional tree long short-term memory network on BiLSTM to obtain substructure information on word sequences and syntactic dependency trees. Liberature [27] used a new global loss function based on previous work. Yith et al. [28] proposed a joint model based on a linear programming formulation that uses the optimal results of subtasks and obtains a global optimal solution. Kate et al. [29] utilize a real-world pyramid structure to model entity and relation information and re-encode the possible entity and relation information in a sentence, so the number of nodes that need to be labeled is greatly reduced. From the labeling strategy perspective, Zheng et al. [30] considered transforming the problem into a single-sequence labeling problem, using an end-to-end network structure [31] to directly extract entity-relation triples. For the relationship overlap problem, Bekoulis et al. [32] continue to propose shared parameters and combine BiLSTM with CRF to propose a multi-head based joint extraction method.

With the application of deep learning in the supervised field, the use of word vectors and word vectors to replace entity feature vectors, and the use of neural network models to extract sentence vectors for classification can solve this problem well. Zeng et al. [33] first proposed the combination of deep learning and remote supervision for entity knowledge extraction, and proposed a PNN model based on the convolutional neural network model. Aiming at the problem of noise introduction that may exist in remote supervision, Ji et al. [34] used a multi-instance method, regarded entity pairs as bags, and selected the indicator with the highest semantic relationship probability as the entity in all sentences containing the same entity pair. The right semantic indicator. The attention mechanism based on Zeng [35] to ensure full utilization of the in-package information while reducing the influence of noise.

## 3 Knowledge extraction method of scientific

## and technological text based on word mixing and GRU (MBGAB)

this paper proposes a knowledge extraction method for scientific and technical texts (MBGAB) is based on word mixing and GRU. A GRU-based end-to-end model is used to generate column sequences for scientific and technological resource text, a bidirectional GRU encodes the input sentence and a GRU decoding layer with bias loss, and finally an objective function with bias weight is used to enhance Relevance of entity tags of scientific and technological resources and reducing the influence of useless tags.

In order to avoid boundary segmentation errors to the greatest extent, the method of word labeling is selected, that is, the input is performed with words as the basic unit. However, in Chinese, it is difficult to store effective semantic information with simple word Embedding. Therefore, in order to integrate semantic information more effectively, a word mixing method is designed. First, input a text sequence in word units, and get a word vector sequence after a word Embedding layer; then segment the text into words and extract the corresponding word vector through a pre-trained Word2Vec model, in order to get the word aligned with the word vector. Vector sequence, the word vector of each word can be repeated as many times as "word number of words"; after obtaining the aligned word vector sequence, the word vector sequence is transformed into the same dimension as the word vector through a matrix, and the two are relative to each other. add. The word mixture vector mapping formula is:

$$w_i = w_{character} + w_{word} \quad (1)$$

Among them, $w_{character}$ represents a single word vector, $w_{word}$ represents a word vector, and a mixed word vector is the sum of the two. The whole process is shown in Figure 1:

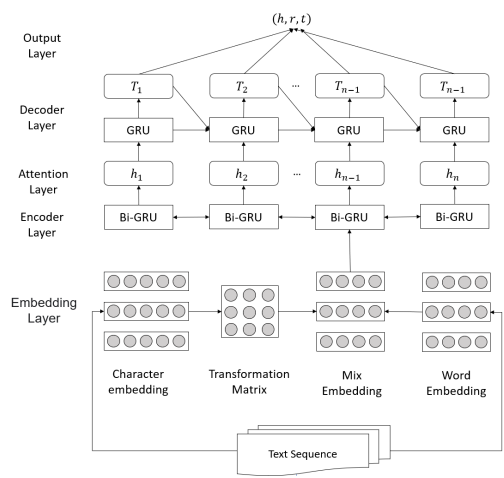

Figure 1. Structure diagram of knowledge extraction method based on word mixing and GRU technology text

After obtaining the text word mixture vector of scientific and technological resources, the word sequence of a sentence can be expressed as $Sen = [w_1, w_2, \ldots, w_n]$, which $w_i$ indicates the $i$th Chinese character in the sentence, and $n$ indicates that the sentence is composed $n$of Chinese characters. For a single Chinese character, $w_i$ its corresponding word embedding vector can be obtained $w_i = [v_{i1}, v_{i2}, \ldots, v_{im}]$ according to the pre-training result, which $m$is the vector dimension of each Chinese character. As shown in the figure, for each time step, according to the previous hidden state $h_{t-1}$and the input word vector $w_t$, the updated hidden state $h_t$ is calculated, and the calculated hidden layer state is $h_t = GRU(w_t, h_{t-1})$as follows:

$$z_t = \gamma(W_z w_t + U_z h_{t-1} + b_z) \quad (2)$$
$$r_t = \gamma(W_r w_t + U_r h_{t-1} + b_r) \quad (3)$$
$$\tilde{h_t} = tanh(W w_t + U h_{t-1} r_t + b) \quad (4)$$
$$h_t = (1 - z_t h_{t-1} + z_t \tilde{h_t}) \quad (5)$$

Among them, $z$it is the upstream data gate and the reset gate $\tilde{h_t}$, respectively. $r$ is the reset hidden unit, $W$, $U$and $b$are parameters that can be learned. For each word $w_t$, the forward GRU layer fuses the above information from the word $w_1$and $w_t$uses it to encode the word $w_t$, denoted by $h_t^f$. Similarly, the backward GRU layer fuses the contextual information from word $w_N$to word $w_t$and uses it to encode the word $w_t$, $h_t^b$ represented by. Finally, splicing them together to represent the encoding information of the t -th word, and the final encoding information is $h_t = [h_t^f + h_t^b]$. After the processing of the Bi - GRU coding layer, the word embedding vector sequence is finally $W = w_1, w_2, \ldots, w_N$converted into a word vector with sentence semantic information $H =$

$\{h_1, h_2, \ldots, h_N\}$.

The attention mechanism can abstract the distance between all words in the sentence as 1, so it can also achieve a good capture effect for the long-distance semantic relationship in the sentence. The input of the self-attention encoding layer comes from the output of the Bi - GRU encoding layer, where the input is $H = h_1, h_2, \ldots, h_N$, and the output is $H^* = h_1^*, h_2^*, \ldots, h_N^*$. First, the input vector is linearly transformed to obtain three vector sequences $Q, K, V$.

The attention calculation formula is:

$$h_i^* = \sum_{j=1}^{N} a_{ij} v_j = \sum_{j=1}^{N} softmax\left(s(q_i, k_j)\right) v_j \quad (6)$$

In this paper, the type of scaled dot product function is used to score attention, and the generated sequence $H^*$ is:

$$H^* = softmax\left(\frac{Q^T K}{\sqrt{d_K}}\right) V \quad (7)$$

The decoding layer is shown in the figure, and the GRU model is used to generate the labeled sequence like the Bi - GRU encoding layer. When labeling words , the input of the decoding layer is: $w_t$ the word vector representation $h_t^*$ obtained from the encoding layer , the previous predicted label representation $T_{t-1}$, and the previous hidden state in the decoding layer, and $h_{t-1}^d$ the predicted label state of the $T_t$ word is obtained after the calculation output $w_t$. The specific formula is as follows:

$$r_t^d = \gamma\left(W_r^d h_t^* + U_r^d h_{t-1}^d + V_r^d T_{t-1} + b_r^d\right) \quad (8)$$

$$z_t^d = \gamma\left(W_z^d h_t^* + U_z^d h_{t-1}^d + V_z^d T_{t-1} + b_z^d\right) \quad (9)$$

$$\widetilde{h_t^d} = tanh\left(W^d r_t^d h_t^* + U^d h_{t-1}^d + V^d T_{t-1} + b^d\right) \quad (10)$$

$$h_t^d = \left(1 - z_t^d\right) h_{t-1}^d + z_t^d \widetilde{h_t^d} \quad (11)$$

$$T_t = tanh\left(W_T h_t^d + b_T^d\right) \quad (12)$$

calculates the entity label probability according to the normalization of the label prediction vector: $T_t$

$$Y_t = W_Y T_t + b_Y \quad (13)$$

$$p_t^i = \frac{exp(Y_t^i)}{\sum_{j=1}^{N_t} exp(Y_t^j)}, i \in 1,\ldots,k \quad (14)$$

Among them, $W_Y$ is a $softmax$ matrix, $Y_t = Y_t^1, \ldots, Y_t^k$ , which represents $Y_t^i$ the prediction relationship distribution of the $N_t$ th tag corresponding to the current word , $i$ represents the total number of tags, and adopts $softmax$ the normalized prediction distribution probability.

Optimization is performed using the RMSprop algorithm by maximizing the log-likelihood function, where the objective function is defined as:

$$L = max \sum_{j=1}^{|D|} \sum_{t=1}^{L_j} \left( \begin{array}{c} log(p_t^j = y_t^j | x_j, \odot)^* I(O) + \\ \alpha^* log(p_t^j = y_t^j | x_j, \odot)^* (1 - I(O)) \end{array} \right) (15)$$

Among them, $|D|$ represents the size of the technical resource text training data set, $L_i$ is the length of the words in the input sentence, $y_t^j$ is $x_j$ the true label $p_t^j$ of the t -th word in the sentence, and is the normalized label probability distribution, $I(O)$ which is used to distinguish the useless label 'O' and the relevant labels that can indicate the extraction result, namely, at that time , $tag =' O' I(O) = 1$; at that time , $tag \neq 'O' I(O) = 0$. In addition, $\alpha$ is the bias weight, a hyperparameter used to control the influence of non-'O' labels, $\alpha$ the larger the label, the greater the influence of the relevant labels in the model. In the serialization and labeling task of this model, in addition to the final triple data information, the sentence also contains other useless information marked by 'O'. These useless information will eventually affect the training results. The bias weight parameter is set to weaken the influence of invalid labels and make the model learn as much as possible in the direction of valid labels.

The overall flow of the algorithm is shown in Table 1.

Table 1 Knowledge extraction algorithm of scientific and technological text based on word mixing and GRU

| Knowledge extraction algorithm of scientific and technological text based on word mixing and GRU |
| --- |
| Input: science and technology resource text $D$, word vector $w_{word}$, dimension $m$ of B i-GRU encoding layer, dimension $n$ of GRU decoding layer , bias weight parameter $\alpha$; |
| Output: Word prediction label status $T_t$. |
| ① fixed word vector $w_{word}$ does not change, and the randomly initialized word vector $w_{character}$ is trained |
| ② get word mix mapping vector $w_i = w_{character} + w_{word}$ |
| ③ After passing through the bidirectional GRU encoding layer, the word embedding vector sequence is $W = w_1, w_2, \ldots, w_N$ converted into a |

word vector with sentence semantic information $H = \{h_1, h_2, .., h_N\}$

④ The input of the self-attention encoding layer comes from the output of the Bi - GRU encoding layer, the input is $H = h_1, h_2, .., h_N$ and the output is $H^* = h_1^*, h_2^*, ..., h_N^*$

⑤ GRU decoding layer obtains the predicted label state of the $T_t$ word by calculating the output $w_t$

# 3 Experimental results and analysis

## 3.1 Dataset

In order to verify the feasibility of the model in the knowledge extraction of scientific and technological resource information of experts and scholars, some Chinese language corpus related to Baidu language and intelligent technology competition and science and technology are used, which contains 19 kinds of relationships, and the data set is divided into 24851 training sets. 6 212 test sets, in which the proportion of labels in the training set and the test set is basically the same to ensure the consistency of the data.

## 3.2 Evaluation indicators

In order to evaluate the effect of the proposed algorithm, this paper uses the precision, recall and F1-Score indicators to evaluate the effect of knowledge extraction.

## 3.3 Comparison algorithm

Use the following algorithms as a comparison to verify the performance of the MBGAB algorithm:

*FCM*: separate entity and knowledge extraction, is a pipeline extraction model

*BiGRU*: remove word mixing embedding, attention mechanism and weight bias

*ME -BiGRU*: removing attention mechanism and weight bias

*ME -BiGRU- SA*: remove weight bias

*BIGRU-SA-Bias*: remove word mixing embedding

*ME-GRU-CRF*: The decoding layer decodes with a conditional random field CRF

*ME -BiGRU- Bias*: remove attention mechanism

The parameters of the experiment are set as follows according to the structure diagram of the algorithm. The input in the coding layer is the word vector generated by the pre-trained Word 2 Vec model, the dimension of the word vector is 300, and the word vector uses a randomly initialized word Embedding layer. During model training, the Word 2 Vec word vector is fixed, and only the transformation matrix and word vector are optimized. The dimension of Bi - GRU encoding layer is 300, the dimension of GRU decoding layer is 600, and the bias weight parameter $\alpha$ is set to 3.

## 3.4 Experiment 1 : Effectiveness of MBGAB Method

In order to verify the effectiveness of the MBGAB method proposed in this paper, this paper adopts the precision rate, recall rate and F1 value as the evaluation indicators of the results. The above- mentioned comparison models were used to conduct comparative experiments, and the experimental results are shown in Table 2.

Table 2 Comparison of experimental results of knowledge extraction

| Contrast algorithm | Accuracy | recall | F1 value |
|---|---|---|---|
| FCM | 53.7 | 33.5 | 41.3 |
| BiGRU | 62.8 | 43.4 | 51.3 |
| ME -BiGRU | 63.6 | 43.8 | 51.9 |
| ME- BiGRU -SA | **67.3** | 46.9 | 55.3 |
| BIGRU-SA-Bias | 64.1 | 47.8 | 55.0 |
| ME-GRU-CRF | 64.6 | 43.9 | 52.3 |
| ME- BiGRU - Bias | 64.1 | 46.5 | 53.9 |
| MBGAB | 65.3 | **4 9.1** | **5 6.1** |

As can be seen from Table 2, because the pipeline-based method executes the two subtasks independently and does not consider the internal correlation between the two subtasks, the efficiency of the pipeline-based extraction model is lower than that of the joint extraction model. Through the comparison between 2 and 3 and 5 and 8, the performance can be improved by 1 percentage point. Considering that the model not only integrates the prior semantic information brought by the pre-trained word vector model, but also retains the flexibility of the word vector, so The effectiveness of this method is proved. Comparing 4 with 8, the model F1 value is improved by 0.8 % , considering that the introduction of paranoid weights enhances the effect of valid entity labels and weakens the influence of invalid labels, so the effect is better. Through the comparison between 3 and 4 and 7 and 8, the F 1 value of the model is increased by 3.4% and 2.2 %, respectively. Considering that the introduction of the self-attention mechanism can effectively capture the long-distance semantic relationship of serialized data,

This improves the performance of the model. By comparing 6 and 8, the decoding layer adopts the GRU model, which is better than the CRF model. Considering that the CRF is good at calculating the joint probability of the label, and the two entity labels associated in the text sentence may be too long, GRU can learn better The long-distance dependencies in the sentence, so the model performance is better.

## 3.5 Experiment 2: End-to-end triple extraction and prediction

In order to further observe the performance of the model in knowledge extraction, end-to-end performance verification of the model is performed. That is, input a sentence, and then output all the triples contained in the sentence. The triplet is $(h, r, t)$ the form, $h$ is the main entity, $t$ is the object entity, $r$ is the relationship between the two entities, and predicate represents the possibility of relationship prediction. Tables 3 and 4 show the triple extraction prediction results of some scientific and technological resource texts.

table 3 Triple Extraction Prediction Results for Text 1

| text 1 | Introduction Lou Tianli, male, researcher, graduated from Zhejiang University of Technology, mainly engaged in college information management | | |
|---|---|---|---|
| main entity $h$ | Lou Yili | | |
| object entity $t$ | Zhejiang University of Technology | | |
| relation | graduated school | Production company | publishing house |
| predicate | 0.8856 | 0.3759 | 0.3238 |

Table 4 Triple extraction prediction results of text 2

| text 2 | Citrus Papilio subspecies Tibet belongs to Animalia, Lepidoptera, Papilioidae | | |
|---|---|---|---|
| main entity $h$ | Citrus Swallowtail subspecies Tibet | | |
| object entity $t$ | Lepidoptera | | |
| relation | target | author | Date of establishment |
| predicate | 0.9628 | 0.0125 | 0.0093 |

It can be seen from Table 3 that the relationship is the most likely to be "graduate school", close to 0.9, while the relationship between "production company" and "publisher" has similar semantics, so there is little difference in the prediction probability. . It can be seen from Table 4 that the possibility of the relationship "mu" is more than 95% . Considering that "mu" can be regarded as a professional vocabulary, and the latter two "author" and "establishment date" are too different in semantics, So the prediction probability is almost zero. Through the above content, the effectiveness of the proposed model in the knowledge extraction task of processing Chinese scientific and technological texts is verified.

## 3.5 Experiment 3: The influence of bias weight on the model

For the bias weight parameter α, when its value is 1, it means that the objective function does not use bias loss, and the same learning weight is used for all labels including the "O" label; when its value is large, it means that it tends to ignore The prediction result of the "O" label, but it may also bring about the problem of decreased accuracy. In order to find a suitable value range, the effect and performance of knowledge extraction under different values are statistically analyzed. The experimental results are shown in Figure 2 .

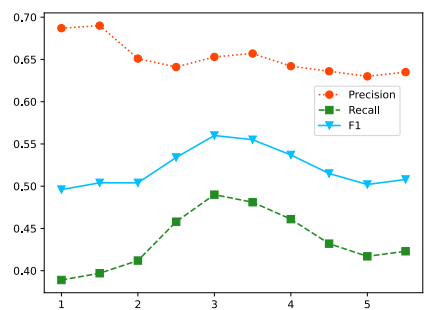

Figure 2 Model extraction effect of bias weights with different values

As the value of the bias weight α increases from 1 to 5, the precision rate of the scientific and technological resource text knowledge extraction model will gradually decrease, while the recall rate and F1 value both increase first and then decrease α; Between 2, the model has higher precision but lower recall, so the F1 value is not ideal. When the α value is around 3, the precision rate and recall rate both achieve a relatively good effect, so the F 1 value is also ideal.

## 4 Conclusion

Aiming at the semantic particularity of Chinese text and the slow convergence of pipeline extraction methods, this paper proposes a knowledge extraction method for scientific and technological texts (MBGAB) based on word mixing and GRU, which effectively improves the knowledge extraction effect for Chinese scientific and technological resource texts. In this paper, a knowledge extraction method of scientific and technological resources based on word mixing and bidirectional GRU is proposed. By adopting an end-to-end model based on GRU to generate the column sequence, the bidirectional GRU encodes the input sentence and a biased The loss of the GRU encoding layer, and finally use an objective function with bias weights to enhance the relevance of entity labels and reduce the influence of useless labels. In order to avoid boundary segmentation errors to the greatest extent , and in order to store more effective semantic information, this paper designs a word-word hybrid vector mapping method. At the same time, combined with the self-attention mechanism, knowledge extraction is carried out for Chinese scientific and technological resource texts. The experimental results show the effectiveness of the proposed method in the knowledge extraction task of scientific and technological resources text data.

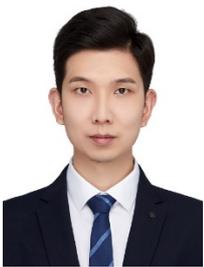

**Suyu Ouyang** was born in 1997, is a Master candidate in Computer Science of Beijing University of Posts and Telecommunications. His research interests include nature language processing, data mining and deep learning .

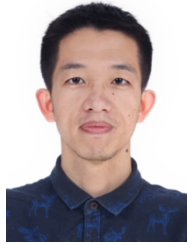

**Yingxia Shao (**corresponding author**)** was born in 1988, male, associate professor, senior member of CCF. The main research areas are large-scale graph analysis, parallel computing framework and knowledge graph analysis .

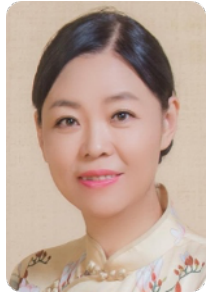

**Junping Du** was born in 1963**.** She is now a professor and Ph.D. tutor at the School of Computer Science and Technology, Beijing University of Posts and Telecommunications. Her research interests include artificial intelligence, machine learning and pattern recognition.

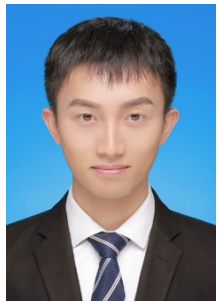

**Ang Li** was born in 1993 , He is currently working toward the Ph.D. degree in Computer Science and Technology at the Beijing University of Posts and Telecommunications, China. His major research interests include information retrieval and data mining.